\documentclass{bmvc2k}

\usepackage{algorithm}
\usepackage{algorithmic}

\usepackage{amsmath}
\usepackage{amssymb}
\usepackage{booktabs}
\usepackage{multirow, adjustbox}
\usepackage{microtype}
\usepackage{enumitem}
\usepackage{mathtools}
\usepackage{array}
\usepackage{float}
\usepackage{pifont}
\newcommand{\xmark}{\ding{55}}%

\DeclareMathOperator*{\argmax}{arg\,max}

\title{EIANet: A Novel Domain Adaptation Approach to Maximize Class Distinction with Neural Collapse Principles}

\addauthor{Zicheng Pan}{zicheng.pan@griffithuni.edu.au}{1}
\addauthor{Xiaohan Yu}{xiaohan.yu@mq.edu.au}{2}
\addauthor{Yongsheng Gao}{yongsheng.gao@griffith.edu.au}{1}

\addinstitution{
 Institute for Integrated and Intelligent Systems,
 Griffith University\\
 Brisbane, Australia
}
\addinstitution{
 School of Computing\\
 Macquarie University\\
 Sydney, Australia
}

\runninghead{Pan, Yu, and Gao}{EIANet for fine-grained source-free domain adaptation}

\begin{document}

\maketitle

\begin{abstract}
Source-free domain adaptation (SFDA) aims to transfer knowledge from a labelled source domain to an unlabelled target domain. A major challenge in SFDA is deriving accurate categorical information for the target domain, especially when sample embeddings from different classes appear similar. This issue is particularly pronounced in fine-grained visual categorization tasks, where inter-class differences are subtle. To overcome this challenge, we introduce a novel ETF-Informed Attention Network (EIANet) to separate class prototypes by utilizing attention and neural collapse principles. More specifically, EIANet employs a simplex Equiangular Tight Frame (ETF) classifier in conjunction with an attention mechanism, facilitating the model to focus on discriminative features and ensuring maximum class prototype separation. This innovative approach effectively enlarges the feature difference between different classes in the latent space by locating salient regions, thereby preventing the misclassification of similar but distinct category samples and providing more accurate categorical information to guide the fine-tuning process on the target domain. Experimental results across four SFDA datasets validate EIANet's state-of-the-art performance. Code is available at: \url{https://github.com/zichengpan/EIANet}.
\end{abstract}

\section{Introduction}
\label{sec:intro}

In recent years, the field of deep learning has seen significant growth in domain adaptation methods, especially for source-free domain adaptation tasks (SFDA). Usually, domain adaptation involves adapting a model trained on a source domain, which has sufficient labelled data, to a target domain where labels are limited or absent. SFDA extends this idea by working under the assumption that the source data is no longer available once the model is trained~\cite{liang2020we}. This approach is particularly useful when dealing with data privacy concerns or when the source data is too large or complex for retraining in the target domain. By not relying on source data, SFDA provides practical and efficient solutions for adapting models, making it well-suited for real-world applications.

\begin{figure}[!t]
\centering

\includegraphics[width=0.7\textwidth]{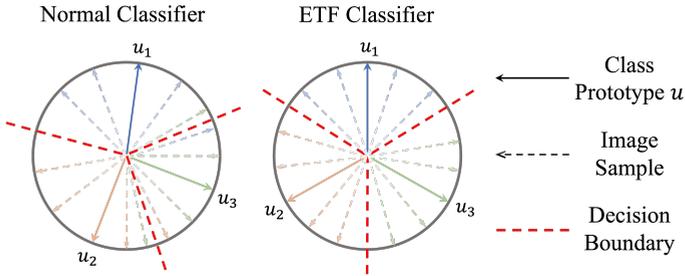}

\caption{Illustration of the proposed ETF classifier compared with normal classifiers. Normal classifiers cannot effectively distinguish samples if two class prototypes are closely related, for example, the samples of class $u_2$ (orange) and class $u_3$ (green) which are close to their corresponding decision boundary can be easily misclassified. The ETF classifier can maximally separate each class's prototype, thereby improving the classification accuracy.}

\label{cherry}
\end{figure}

To adapt a model pre-trained on a labelled source domain to an unlabelled target domain, modern state-of-the-art SFDA methods typically generate categorical information for target domain training. They achieve this either by creating pseudo-labels or by grouping similar feature embeddings in the latent space as the same class~\cite{liang2020we, yang2021exploiting, yang2022attracting, yi2023when}. However, these methods can mistakenly categorize different class samples as the same class when their embeddings appear similar. This problem becomes particularly severe with fine-grained tasks~\cite{wang2020progressive, pan2024pseudo} where differences between classes are small. Moreover, standard SFDA approaches often struggle to identify key features that accurately represent each class. These limitations hinder the effectiveness of the adaptation, especially when distinguishing between closely related categories in the target domain. Addressing these issues requires an approach that can recognize and emphasize the salient regions of the image as well as mitigate the generation of false categorical information, ensuring more accurate classification and better adaptation to the new domain.

To address these challenges, we introduce a novel ETF-Informed Attention Network (EIANet) inspired by the neural collapse phenomenon observed in deep learning~\cite{papyan2020prevalence}. Neural collapse refers to the empirical observation that, in the later stages of training a deep network, the feature vectors of samples within the same class tend to converge to a class-specific mean (prototype), forming vertices of a simplex Equiangular Tight Frame (ETF). The classifier weights also converge to the same ETF to simplify the classification task. EIANet leverages this phenomenon, integrating an ETF classifier with an attention mechanism to enhance feature extraction and class prototype separation. By exploiting the principles of neural collapse, the ETF classifier is specially designed to maximize the distinction between class prototypes. This is achieved by constructing a classifier whose decision boundaries are equidistant to each centroid, ensuring maximal inter-class distinction. As shown in Fig.~\ref{cherry}, normal classifiers rely on class prototype weights that are derived from training. However, these class prototypes can be situated very close to each other, which makes it difficult to distinguish between their corresponding class samples. This difficulty arises because the decision boundaries are too close to the class prototypes. In contrast, the ETF classifier is designed to overcome this limitation. By forcing each class prototype not only distinct but also equidistant from others, it ensures that there is sufficient space to establish distinct decision boundaries. Consequently, this separation leads to better clustering of samples and enhances the classifier's ability to accurately categorize them. 
This approach is particularly advantageous in fine-grained datasets, where standard methods often struggle to differentiate between closely related samples in the unlabelled target domain.
Since the ETF classifier is pre-defined at the beginning and does not get updated during training, we further integrate an attention mechanism to better align the feature with the ETF classifier prototype weights. It makes the model focus on the most discriminative parts of images, helping the network to recognize and emphasize important features that are often overlooked by standard approaches. By doing so, EIANet can effectively reduce the likelihood of utilizing false category information in the target domain, providing a robust and reliable tool for effective knowledge transfer across domains.

The contributions of our work can be summarized as follows:

\begin{itemize}
    \item We proposed a novel ETF-Informed Attention Network (EIANet) which utilizes the principle of neural collapse phenomenon by combining a simplex Equiangular Tight Frame (ETF) classifier for SFDA tasks. We innovate an algorithm that utilizes SVD decomposition to initialize the ETF classifier and satisfy its properties by definition. This integration notably enhances class prototype separation to provide accurate categorical information in the unlabelled target domain for model adaptation.
    
    \item An attention mechanism is integrated into the network to facilitate the model generating image features that align with the pre-defined ETF classifier. Furthermore, the integration of the attention mechanism allows the model to concentrate on the most discriminative features within images, which aids in accurately identifying salient regions and better distinguishing similar samples.

    \item Extensive experiments are conducted on four SFDA datasets to validate the effectiveness of EIANet. Our results demonstrate state-of-the-art performance, highlighting the model's robustness and adaptability across various domain adaptation scenarios.
    
\end{itemize}

\section{Related Works}

\subsection{Preliminaries of Neural Collapse Phenomenon}
Neural collapse describes a phenomenon observed in the terminal phase of training deep neural networks, particularly when the training error rate closes to zero. This phenomenon exhibits a geometric structure in the space of the last-layer features and classifiers. It can be characterized by four distinct properties~\cite{papyan2020prevalence}:

\begin{description}

    \item[NC1 (Variability Collapse):] The within-class sample features from the last layer have nearly zero variation, meaning that the image embeddings collapse to their corresponding class means.
    \item[NC2 (Convergence to Simplex ETF):] The centred within-class means align to form vertices of a simplex Equiangular Tight Frame (ETF), which is a configuration of vectors with equal length and angles and maximal pairwise distances within the constraints of the previous properties.
    \item[NC3 (Convergence to Self-Duality):] Within-class means and the classifier weights, converge to a symmetric structure, which means that the classifier weights converge to the same simplex ETF.
    \item[NC4 (Simplification to Nearest Class Center):] The classification process simplifies to select the class whose mean is closest to the given embedding in the feature space.
\end{description}

After establishing the properties of neural collapse, the simplex ETF structure based on these properties can be defined as follows.

\textbf{Simplex Equiangular Tight Frame (ETF).}
    A simplex Equiangular Tight Frame (ETF) in $\mathbb{R}^d$ is a matrix composed of $K$ vectors that satisfies the following condition~\cite{yang2023neural}:
    \begin{equation}
        \mathbf{E} = \sqrt{\frac{K}{K-1}} \mathbf{U} \left( \mathbf{I}_K - \frac{1}{K} \mathbf{1}_K \mathbf{1}_K^T \right),
    \label{eq1}
    \end{equation}
    where $\mathbf{E} = [\mathbf{e}_1, \ldots, \mathbf{e}_K] \in \mathbb{R}^{d \times K}$, $\mathbf{U} \in \mathbb{R}^{d \times K}$ is a rotation matrix with $\mathbf{U}^T\mathbf{U} = \mathbf{I}_K$, $\mathbf{I}_K$ is the identity matrix, and $\mathbf{1}_K$ is an all-ones vector. The columns of $\mathbf{E}$ have equal $\ell_2$ norm, and any pair of distinct columns has an inner product of $-\frac{1}{K-1}$.

\section{Methodology}
\subsection{Source-Free Domain Adaptation Formulation}

In the SFDA framework, following the commonly used protocols~\cite{liang2020we, yang2021exploiting}, we consider a source domain dataset $\mathcal{D}_s = \{(\mathit{x}_s^i, \mathit{y}_s^i)\}_{i=1}^{N}$, where $N$ is the total number of classes, $\mathit{x}_s^i$ and $\mathit{y}_s^i$ denote the $i$-th class samples and its corresponding label in the source domain respectively. A model $f_s$ is initially trained on $\mathcal{D}_s$ under supervised conditions. After the training is done, the task involves adapting the pre-trained $f_s$ to a target domain dataset $\mathcal{D}_t = \{\mathit{x}_t^i\}_{i=1}^{N}$, where labels $\mathit{y}_t^i$ are absent in $\mathcal{D}_t$. The adaptation aims to leverage the knowledge from $f_s$ and apply it to $\mathcal{D}_t$. The source domain and target domain data share the same label space, and the adapted model's performance is evaluated on the labelled target domain dataset after training. 
In this work, we introduce the ETF-Informed Attention Network (EIANet). It utilizes the neural collapse phenomena and leverages simplex Equiangular Tight Frame (ETF) principles to maximally separate each class's prototype. Details of the ETF classifier, attention mechanism, and our proposed EIANet training processes are introduced in the following sections.

\subsection{ETF Classifier Construction}

Given a classification problem with a feature dimension of $d$ and $K$ classes, inspired by~\cite{yang2023neural}, the simplex Equiangular Tight Frame (ETF) classifier is constructed as follows:

\begin{enumerate}
    \item \textbf{Parameter Initialization:} The classifier is defined by a matrix $\mathbf{E} \in \mathbb{R}^{d \times K}$, where each column of $\mathbf{E}$ represents a prototype vector for each class. The number of features $d$ should be greater than or equal to the number of classes $K$.
    
    \item \textbf{Generation of Rotation Matrix $\mathbf{U}$:} Construct a $d \times K$ rotation matrix $\mathbf{U}$ by performing Singular Value Decomposition (SVD) on a randomly generated $d \times K$ matrix. The rotation matrix satisfies the orthogonal property, $\mathbf{U}^T \mathbf{U} = \mathbf{I}_K$, where $\mathbf{I}_K$ is the $K \times K$ identity matrix.
    
    \item \textbf{Construction of ETF Matrix $\mathbf{E}$:} The ETF matrix is computed using Eq.~\ref{eq1} where $\mathbf{1}_K$ is a $K \times K$ matrix with all elements equal to 1. This step ensures that the columns of $\mathbf{E}$ are equiangular to each other.
    
    \item \textbf{Classifier Integration:} The matrix $\mathbf{E}$ is integrated into the neural network architecture as a linear classifier layer. The parameters of $\mathbf{E}$ are kept fixed (non-trainable) during the training process to maintain the properties of the ETF structure.
\end{enumerate}

The constructed ETF classifier is validated to satisfy the following properties: 
1) Each column vector of $\mathbf{E}$ has the same length of $\sqrt{\frac{K}{K-1}}$;
2) The inner product between any two distinct column vectors of $\mathbf{E}$ is constant and equal to $-\frac{1}{K-1}$.
These properties confirm the equiangular nature of the classifier. Each column of the classifier is treated as a class prototype, making it maximally distinguish each class's prototype to achieve accurate clustering.

Define each class prototype as \( u^{i} \), which corresponds to the \( i \)-th column of the matrix \( \mathbf{E} \). The classification of a sample \( x \) is determined by selecting the class prototype that has the highest similarity with \( x \). This prediction process is formulated as:

\begin{equation}
pred = \underset{i \in \{1, 2, \ldots, N\}}{\argmax} \left[ \cos(x, u^{i}) \right],
\end{equation}

\noindent where \( \cos \) indicates cosine similarity and \( i \) corresponds to the \( i \)-th class. The sample \( x \) is assigned to the class whose prototype is most similar to \( x \) in terms of cosine similarity.

\begin{figure*}[!t]
\centering

\includegraphics[width=\textwidth]{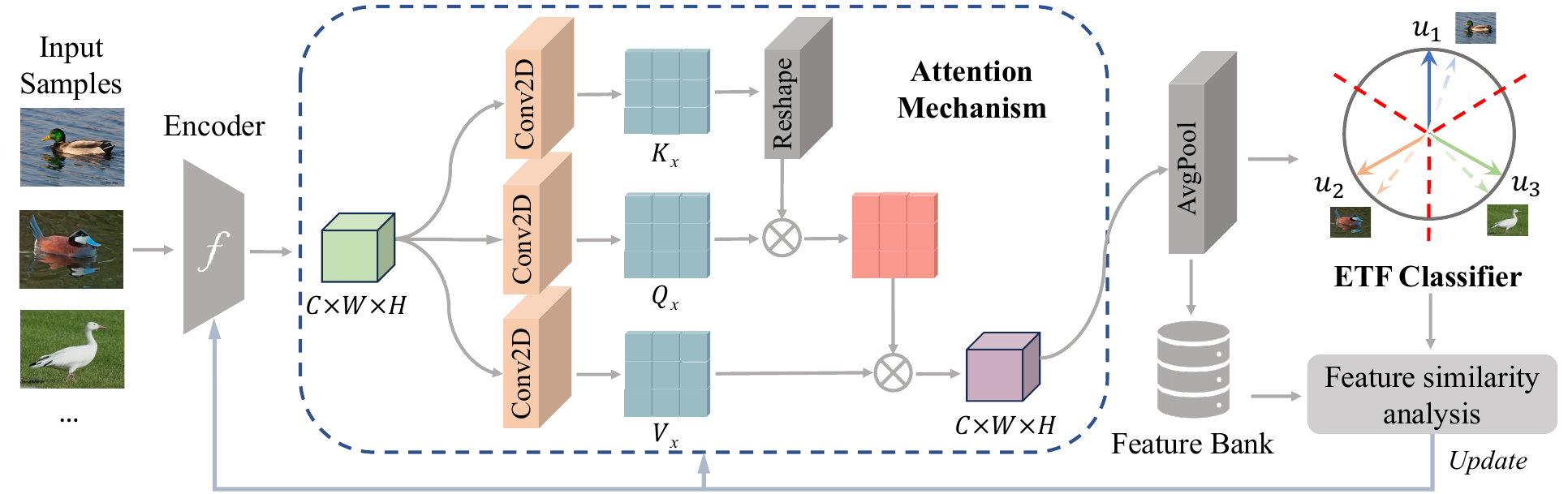}

\caption{The clustering process of the proposed EIANet during target domain adaptation. The ETF classifier is set up at the beginning before source domain training and remains consistent. Input images first undergo feature extraction via an image encoder with the attention mechanism highlighting salient regions. The refined data is then classified based on the proximity of the image's features to the distinct class prototypes in the ETF classifier.}

\label{network}
\end{figure*}

\subsection{Attention Mechanism}

To enhance the model's ability to focus on more discriminative features and facilitate the effectiveness of the fixed simplex ETF classifier, we integrate an attention module before the average pooling layer of the network. It can dynamically reweight the input feature map, allowing the model to focus on the most relevant parts of the input. In this work, we adopt self-attention which is widely used in the Transformer architecture~\cite{vaswani2017attention}. As shown in Fig.~\ref{network}, the image embeddings from the encoder are getting refined by the attention module, which can effectively locate discriminative regions and better align with their corresponding class prototypes in the ETF classifier. The self-attention mechanism operates by transforming the input feature map into query ($Q_x$), key ($K_x$), and value ($V_x$) representations through convolutional layers. The query and key representations are generated in a reduced dimensionality space, while the value representation maintains the original dimensions. The core of the self-attention mechanism is the computation of the attention map, which is formulated as follows:

\begin{equation}
\begin{aligned}
& \mathsf{Attention}(Q_x, K_x, V_x) = \mathsf{softmax}\left(\frac{Q_x K_x^{T}}{\sqrt{G_x}}\right)V_x,
\end{aligned}
    \label{eq3}
\end{equation}

\noindent where $G_x$ is a scaling factor derived from the dimensionality of the key $K_x$. The attention map provides a set of weights that indicate the importance of each spatial location in the feature map. The final output of the self-attention module is a combination of the input feature map and the attention-weighted value map. By incorporating this self-attention mechanism, our network is able to effectively move the sample embeddings toward their corresponding class prototypes by salient features and enhance the overall classification performance.

\subsection{Training Procedures on Source and Target Domains}

For the training on the source and target domain, we follow the training strategy similar to conventional approaches~\cite{liang2020we,chu2022denoised,yang2021exploiting,yang2022attracting}, with the key distinction being the integration of our predefined ETF classifier and the attention mechanism. The model is first trained on a well-labelled source domain dataset using cross-entropy loss with label smoothing and the ETF classifier. 
The pre-trained source model $f_s$ will be adapted to the target domain as $f_t$, which incorporates the feature similarity analysis via similarity loss and diversity loss by constructing feature banks as in~\cite{yang2021exploiting,yang2022attracting,yang2021generalized}. More specifically, we first construct the feature bank $B_t$ for all sample embeddings in the target domain and the feature bank $B_m$ for the embeddings in the training mini-batch. Then $M$ nearest neighbour samples, denoted as $ \mathit{z}_{x}$, are selected from the $B_t$ as positive samples for each training sample. The positive samples have similar features and most likely belong to the same category. We adopt Kullback–Leibler divergence ($KL$) as the similarity loss $\textit{L}_{sim}$ to pull the training sample embedding $f_t(x)$ towards the average embedding of nearby $M$ samples, which can be formulated as:

\begin{equation}
\begin{aligned}
& \textit{L}_{sim} = KL(f_t(x), \frac{1}{M}\sum_{j=1}^{M} f_t(\mathit{z}^{j}_{x})),
\end{aligned}
\label{eq4}
\end{equation}

\noindent where $z_x$ denoted nearby samples around $x$. In addition, a study has shown that the features in $B_t$ have a higher chance of containing similar features than features in $B_m$~\cite{yang2022attracting}. Since the similarity loss $\textit{L}_{sim}$ has already considered the positive cases from the whole target domain's perspective, we can simply treat other samples in the mini-batch as negative samples to input $x$ and enlarge their diversity via:

\begin{equation}
\begin{aligned}
& \textit{L}_{div} = \frac{1}{P}\sum_{p=1}^{P} dist(f_t(x),f_t(\mathit{s}^{p}_{x})),
\end{aligned}
\label{eq5}
\end{equation}

\noindent where $P$ is the number of negative samples, $s$ is other samples in the mini-batch, and $dist$ represents the Euclidean distance between two sample representations. Finally, the fine-tuning loss $\textit{L}_{t}$ for the target domain model becomes:

\begin{equation}
\begin{aligned}
& \textit{L}_{t} = \textit{L}_{sim} + \alpha\times\textit{L}_{div},
\end{aligned}
\label{eq6}
\end{equation}

\noindent where $\alpha$ is the weight assigned to the diversity loss. Note that the proposed simplex ETF classifier weights are set up initially before source domain training and kept frozen during the whole training and adaptation process. Only the backbone network and the attention module are trainable. This approach is specifically designed to maximally separate the prototypes of each class, thus enhancing the discriminative power of the model.

\section{Experimental Results}

\subsection{Datasets}
The proposed EIANet is evaluated on traditional SFDA datasets including Office-Home~\cite{venkateswara2017deep} and Office-31~\cite{saenko2010adapting} which follows the common settings as~\cite{liang2020we, yi2023when, chhabra2023generative}. Office-Home comprises four distinct domains: \textbf{A}rtistic images, \textbf{C}lipart, \textbf{P}roduct images, and \textbf{R}eal-World images, each containing a diverse array of everyday office objects that belong to 65 classes. Office-31 has 31 categories, which also cover a broad range of office-related objects. It contains 4,625 images in total which consists of three domains: \textbf{A}mazon, \textbf{W}ebcam, and \textbf{D}SLR. 

In addition, we further conduct experiments to evaluate the proposed method on fine-grained domain adaptation datasets, e.g., CUB-Paintings~\cite{wang2020progressive} and Birds-31. CUB-Paintings consists of two domains (CUB-\textbf{P}ainting~\cite{wang2020progressive} and \textbf{C}UB-200-2011~\cite{WelinderEtal2010}) with each containing 200 categories. CUB-Painting consists of artistic representations of 3,047 images across 200 bird species. CUB-200-2011 contains realistic images of the same species with 11,788 samples in total. Birds-31 has three domains (\textbf{C}UB-200-2011~\cite{WelinderEtal2010}, \textbf{N}ABirds~\cite{van2015building}, \textbf{i}Naturalist2017~\cite{van2018inaturalist}), where 31 common bird species from these domains are used.

\begin{table*}[h]
\centering

\caption{Benchmark performance (\%) on the Office-Home dataset.}

\scalebox{0.85}{
\begin{tabular}{p{62pt} |  p{15pt}<{\centering} p{15pt}<{\centering}  p{15pt}<{\centering}  p{15pt}<{\centering}  p{15pt}<{\centering}  p{15pt}<{\centering}  p{15pt}<{\centering}  p{15pt}<{\centering}  p{15pt}<{\centering} p{15pt}<{\centering} p{15pt}<{\centering} p{18pt}<{\centering} |p{15pt}<{\centering}}
\toprule
    
Methods & A$\rightarrow$C & A$\rightarrow$P & A$\rightarrow$R & C$\rightarrow$A & C$\rightarrow$P & C$\rightarrow$R & P$\rightarrow$A & P$\rightarrow$C & P$\rightarrow$R & R$\rightarrow$A & R$\rightarrow$C & R$\rightarrow$P & Avg. \\
\midrule

ResNet-50 & 34.9 &50.0 &58.0& 37.4& 41.9& 46.2& 38.5& 31.2& 60.4& 53.9& 41.2& 59.9& 46.1\\
SHOT  & 57.1 & 78.1 & 81.5 & 68.0 & 78.2 & 78.1 & 67.4 & 54.9 & 82.2 & 73.3 & 58.8 & 84.3 & 71.8\\
G-SFDA  & 57.9& 78.6 &81.0& 66.7& 77.2 &77.2& 65.6& 56.0& 82.2& 72.0& 57.8& 83.4 &71.3\\
NRC & 57.7& 80.3& 82.0& 68.1& 79.8& 78.6& 65.3 &56.4& 83.0 &71.0& 58.6& 85.6& 72.2\\
U-SFAN+  &  57.8 &77.8 &81.6 &67.9 &77.3 &79.2& 67.2 &54.7& 81.2& 73.3& 60.3 &83.9& 71.9\\
D-MCD &59.4 &78.9& 80.2& 67.2& 79.3& 78.6& 65.3 &55.6& 82.2& 73.3 &62.8& 83.9& 72.2\\
CoWA-JMDS  &  56.9& 78.4& 81.0& 69.1& 80.0 &79.9& 67.7 &57.2& 82.4& 72.8 &60.5& 84.5 &72.5 \\
PLUE &49.1& 73.5& 78.2& 62.9 &73.5& 74.5 &62.2& 48.3 &78.6& 68.6& 51.8& 81.5 &66.9 \\
GAP  &55.4& 73.4 &80.8& 67.2& 75.5 &78.3& 65.5 &54.0 &82.4 &74.3& 59.4& 84.0 &70.8\\
NRC+ELR & 58.4 &78.7 &81.5 &69.2 &79.5& 79.3 &66.3& 58.0 &82.6 &73.4& 59.8 &85.1 &72.6\\
SFADA & 56.1& 78.0& 81.6 &68.5 &79.5& 78.5 &67.8& 56.0 &82.3 &73.6 &57.8& 83.0& 71.9\\
\midrule

\textbf{EIANet} & 59.7& 80.6& 82.8 & 69.6 & 81.3 & 79.2& 67.6& 57.5 & 82.0& 74.0 & 60.4 & 85.7& \textbf{73.4}
\\ \bottomrule

\end{tabular}
}
\label{tab:home}
\end{table*}

\subsection{Implementation Details}

Following the common training settings~\cite{liang2020we, yang2021exploiting, yi2023when}, we adopt the same image pre-processing procedures and employ the ResNet-50 architecture~\cite{he2016deep} with the ImageNet~\cite{deng2009imagenet} pre-trained weights as the backbone for all datasets. SGD is used as the optimizer with a momentum of 0.9 and weight decay of $5e^{-4}$. The neighbour number $M$ is optimized to 4 and the hyperparameter \(\alpha\) is defined as: \( \alpha = 0.1\times M\). Notably, the proposed ETF classifier at the end of the network is kept fixed throughout the training process. More detailed configurations of experiments as well as datasets can be found in the released code repository. All experiments are conducted with the PyTorch library~\cite{NEURIPS2019_9015} on multiple RTX A5000 GPUs.

\subsection{Compared with State-of-the-art Methods}
To demonstrate the effectiveness of the proposed method, we compared it with other state-of-the-art methods including ResNet-50~\cite{he2016deep}, SAFN~\cite{xu2019larger}, BCDM~\cite{li2021bi}, MCD~\cite{saito2018maximum}, CDAN~\cite{long2018conditional}, CDAN+BSP~\cite{chen2019transferability}, PAN~\cite{wang2020progressive}, SHOT~\cite{liang2020we}, G-SFDA~\cite{yang2021generalized}, NRC~\cite{yang2021exploiting}, U-SFAN+~\cite{roy2022uncertainty}, D-MCD~\cite{chu2022denoised}, CoWA-JMDS~\cite{lee2022confidence}, PLUE~\cite{litrico2023guiding}, GAP~\cite{chhabra2023generative}, NRC+ELR~\cite{yi2023when}, and SFADA~\cite{he2024source}. Some benchmark results are obtained from~\cite{tang2023source}. Top-1 accuracy is recorded for each task and the best average performance (Avg.) is highlighted in bold. 

\textbf{SFDA on conventional datasets.}
The experimental results conducted on common SFDA datasets (Office-Home and Office-31) are reported in Tables~\ref{tab:home} and~\ref{office31}. Our proposed EIANet consistently achieved the best average performance across both datasets. This outcome demonstrates its great adaptability and effectiveness in common domain adaptation scenarios. These results reinforce the potential of EIANet as a leading solution in the field of SFDA, setting a new benchmark for future research and applications.

\begin{table}[!h]
\caption{Benchmark performance (\%) on the Office-31 dataset.}

\centering
\scalebox{0.85}{

\begin{tabular}{l|cccccc|c}
\toprule
Methods &
A$\rightarrow$D & A$\rightarrow$W & D$\rightarrow$A & D$\rightarrow$W & W$\rightarrow$A & W$\rightarrow$D & Avg. \\

\midrule
ResNet-50& 68.9& 68.4& 62.5 &96.7& 60.7& 99.3& 76.1 \\
BCDM  & 93.8& 95.4& 73.1& 98.6& 73.0& 100.0& 89.0\\

SHOT  & 94.0 & 90.1 & 74.7 & 98.4 & 74.3 & 99.9 & 88.6\\

NRC & 96.0 &90.8 &75.3 &99.0 &75.0 &100.0 &89.4\\
U-SFAN+ & 94.2& 92.8 &74.6& 98.0 &74.4 &99.0 &88.8 \\
D-MCD &94.1& 93.5 &76.4& 98.8 &76.4 &100.0 &89.9\\
PLUE & 89.2 &88.4& 72.8 &97.1 &69.6 &97.9& 85.8\\
GAP& 90.6 &90.9 &74.5 &98.7 &73.9 &99.8 &88.1\\
NRC+ELR& 93.8 &93.3 &76.2 &98.0 &76.9 &100.0& 89.6\\
SFADA& 94.8 &92.0&76.5& 97.6& 75.7& 99.8 & 89.4\\
\midrule
\textbf{EIANet} & 96.4 & 94.5& 76.6& 99.0& 75.7& 100.0& \textbf{90.4}
\\
\bottomrule
\end{tabular}
}
\label{office31}
\end{table}

\textbf{SFDA on fine-grained datasets.}
To validate that the proposed EIANet has superior domain adaptation ability and can distinguish similar objects, we conduct experiments on more challenging fine-grained domain adaptation datasets, CUB-Paintings and Birds-31. The benchmark comparison results are recorded in Table~\ref{combined_benchmarks}. 

\begin{table}[ht]
\centering
\caption{Benchmark performance (\%) on the fine-grained (CUB-Paintings and Birds-31) datasets with the ResNet-50 backbone.}
\scalebox{0.85}{
\begin{tabular}{l|ccc|ccccccc}
\toprule
\multirow{2}{*}{Methods} & \multicolumn{3}{c|}{CUB-Paintings} & \multicolumn{7}{c}{Birds-31} \\
& C$\rightarrow$P & P$\rightarrow$C & Avg. & C$\rightarrow$I & I$\rightarrow$C & I$\rightarrow$N & N$\rightarrow$I & C$\rightarrow$N & N$\rightarrow$C & Avg. \\
\midrule
ResNet-50 & 47.88 & 36.62 & 42.25 & 64.25 & 87.19 & 82.46 & 71.08 & 79.92 & 89.96 & 79.14 \\
MCD & 63.40 & 43.63 & 53.52 & 66.43 & 88.02 & 85.57 & 73.06 & 82.37 & 90.99 & 81.07 \\
CDAN & 63.18 & 45.42 & 54.30 & 68.67 & 89.74 & 86.17 & 73.80 & 83.18 & 91.56 & 82.18 \\
CDAN+BSP & 63.27 & 46.62 & 54.95 & 68.64 & 89.71 & 85.72 & 74.11 & 83.22 & 91.42 & 82.13 \\
SAFN & 61.38 & 48.86 & 55.12 & 65.23 & 90.18 & 84.71 & 73.00 & 81.65 & 91.47 & 81.08 \\
PAN & 67.40 & 50.92 & 59.16 & 69.79 & 90.46 & 88.10 & 75.03 & 84.19 & 92.51 & 83.34 \\
\midrule
\textbf{EIANet} &  69.61 & 53.33 & \textbf{61.47} & 73.22 & 92.37 & 91.27 & 79.14 & 86.51 & 94.75 & \textbf{86.21} \\
\bottomrule
\end{tabular}
}
\label{combined_benchmarks}
\end{table}

From the table, it is observed that the proposed EIANet significantly enhances prediction accuracy compared to other benchmark methods, achieving notable average performance gains of 2.31\% and 2.87\% over the second-best method (PAN) on the two datasets respectively. These improvements underscore EIANet's robust identification and adaptation capabilities in complex and challenging fine-grained domain adaptation tasks characterized by high similarity among objects. EIANet's effectiveness in these scenarios is attributed to its ability to maximally separate class prototypes and more effectively cluster similar objects.

\subsection{Ablation Study of Attention and ETF Classifier}
To evaluate the effectiveness of components in the proposed EIANet, we conducted ablation experiments on the Office-31 and CUB-Paintings datasets. For the Office-31 dataset, we assessed the performance of the baseline model and compared it to the implementations with an attention mechanism (without ETF) and with both an attention mechanism and the ETF classifier (with ETF). Furthermore, we analyzed the influence of the attention mechanism and the ETF classifier on the CUB-Paintings dataset by testing different combinations of their presence. The results of these experiments are presented in Figure~\ref{radar} and in Table~\ref{cub_ablation}.

\begin{figure}[!ht]
    \centering
    \begin{minipage}{0.48\textwidth}
        \centering
        \includegraphics[width=\linewidth]{figures/ablation_radar.pdf} 
        \caption{Components analysis on Office-31.}
        \label{radar}
    \end{minipage}\hfill
    \begin{minipage}{0.48\textwidth}
        \centering
        \begin{table}[H]
        \centering
        \caption{Component influence analysis on the CUB-Paintings dataset with/without attention mechanism and ETF classifier.}
        \scalebox{0.85}{
        \begin{tabular}{p{28pt}<{\centering} p{25pt}<{\centering} | p{26pt}<{\centering}  p{26pt}<{\centering} | p{30pt}<{\centering}}
            \toprule
            Attention & ETF & C$\rightarrow$P & P$\rightarrow$C & Avg.\\ 
            \midrule
            \xmark & \xmark & 66.92 & 50.35 & 58.64 \\
            \xmark & \checkmark & 68.49 & 52.68 & 60.58 \\
            \checkmark & \xmark & 67.90 & 51.94 & 59.92 \\
            \checkmark & \checkmark &  69.61 & 53.33 & 61.47  \\
            \bottomrule
        \end{tabular}
        }
        \label{cub_ablation}
        \end{table}
    \end{minipage}
\end{figure}

From the results, it is shown that for the Office-31 dataset, the attention mechanism has shown improvements over the baseline method. However, in fine-grained tasks such as CUB-Paintings, the attention mechanism alone struggles to distinguish between different classes effectively. Our proposed ETF classifier acts as a guiding force for achieving maximum class separation in the latent space. This guidance allows the attention mechanism to focus more precisely on the salient regions critical for distinguishing between classes. Consequently, the combination of the ETF classifier and the attention mechanism (EIANet) significantly enhances the model's ability to handle complex, fine-grained domain adaptation tasks and achieve optimal performance.

\section{Conclusion}

In this paper, we proposed a novel ETF-Informed Attention Network (EIANet) to address the challenges of source-free domain adaptation (SFDA) tasks. EIANet innovatively combines a well-designed simplex Equiangular Tight Frame (ETF) classifier with an attention mechanism, leveraging the neural collapse phenomenon to enhance class prototype separation and distinction within the classifier. This innovative approach enables the model to more effectively identify salient regions in images. Furthermore, EIANet facilitates maximal separation of class prototypes, allowing for more precise differentiation and clustering of unlabelled target domain data. This capability ensures the provision of accurate categorical information, which is essential for fine-tuning the model in the target domain. Our approach demonstrates a notable improvement in performance across both traditional and fine-grained domain adaptation datasets, which offers valuable insights for future advancements in the field.

\section{Acknowledgement}

This work was supported in part by the Australian Research Council under the Industrial Transformation Research Hub Grant IH180100002 and the Discovery Grant DP180100958.

\bibliography{egbib}
\end{document}